\documentclass[10pt,twocolumn,letterpaper]{article}
\usepackage{cvpr}
\usepackage{times}
\usepackage{epsfig}
\usepackage{graphicx}
\usepackage{amsmath}
\usepackage{amssymb}
\usepackage{soul}
\usepackage{floatrow}
\usepackage{multirow}
\usepackage{placeins}
\usepackage{booktabs}
\usepackage{scrextend}
\usepackage{soul}

\usepackage[symbol]{footmisc}

\newcommand{\mycomment}[1]{}

\usepackage{bm}
\usepackage{blindtext}
\usepackage{caption}

\usepackage[pagebackref=true,breaklinks=true,unicode=true,letterpaper=true,colorlinks,bookmarks=false]{hyperref}
\usepackage{cleveref}

\cvprfinalcopy

\newfloatcommand{capbtabbox}{table}[][\FBwidth]

\newcommand{\cmt}[1]{\ignorespaces}

\vbadness=\maxdimen

\makeatletter
\renewcommand{\paragraph}{%
  \@startsection{paragraph}{4}%
  {\z@}{0.5em}{-1em}%
  {\normalfont\normalsize\bfseries}%
}
\makeatother

\def\cvprPaperID{0033}

\ifcvprfinal\fi

\title{Slim DensePose: Thrifty Learning from Sparse Annotations and Motion Cues}
\author{\noindent Natalia Neverova$\vphantom{a}^{1}$\hspace*{0.4cm} James Thewlis$\vphantom{a}^{2*}$\hspace*{0.4cm} R{\i}za Alp G\"uler$\vphantom{a}^{3}$\hspace*{0.4cm} Iasonas Kokkinos$\vphantom{a}^{3*}$\hspace*{0.4cm} Andrea Vedaldi$\vphantom{a}^{1}$\\
$\vphantom{a}^{1}$Facebook AI Research\hspace*{1cm} $\vphantom{a}^{2}$University of Oxford\hspace*{1cm} $\vphantom{a}^{3}$Ariel AI
}

\begin{document}

\twocolumn[{%
\maketitle
\vspace*{-2mm}
    \includegraphics[width=1\textwidth,trim={0 0 0 0},clip]{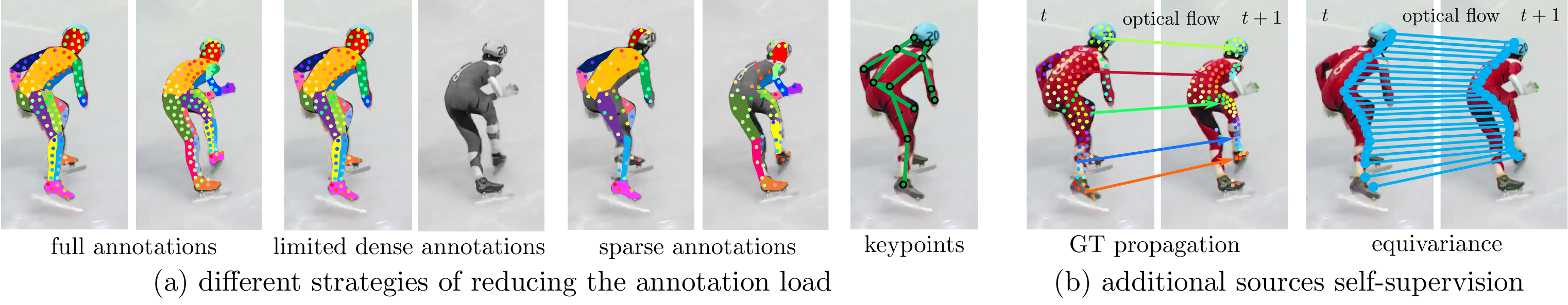}
\captionof{figure}{The goal of this work is to discover effective and cost-efficient data annotation strategies for the task of learning dense correspondences in the wild (DensePose). We significantly reduce the annotation effort by exploiting (a) sparse subsets of the DensePose labels augmented with cheaper kinds of annotations, such as object masks or keypoints, and (b) temporal information in videos to propagate ground truth  and enforce dense spatio-temporal equivariance constraints.}\label{f:splash}
}]\vspace*{1mm}

%we could say the first three are ``sparser'', as in sparse in data, sparse in space and that time makes it easy to get denser supervision without extra annotation effort (an in 'denser in data' - GT propagation,  and 'dense in space' - equivariance constraint)
%well, let's see how this would look like I guess
%btw, why equivariance = denser in space as opposed to denser in data?
%I mean, it also allows to use more data points (more images)
%in addition to using more/denser points
%sure - maybe let's say dense in both
%the main thing is that equivariance is applicable over the whole image domain 

\begin{abstract}
DensePose supersedes traditional landmark detectors by densely mapping image pixels to body surface coordinates.
This power, however, comes at a greatly increased annotation time, as supervising the model requires to manually label hundreds of points per pose instance.
In this work, we thus seek methods to significantly slim down the DensePose annotations, proposing more efficient data collection strategies.
In particular, we demonstrate that if annotations are collected in video frames, their efficacy can be multiplied for free by using motion cues.
To explore this idea, we introduce DensePose-Track, a dataset of videos where selected frames are annotated in the traditional DensePose manner.
Then, building on geometric properties of the DensePose mapping, we use the video dynamic to propagate ground-truth annotations in time as well as to learn from Siamese equivariance constraints.
Having performed exhaustive empirical evaluation of various data annotation and learning strategies, we demonstrate that doing so can deliver significantly improved pose estimation results over strong baselines.
However, despite what is suggested by some recent works, we show that merely synthesizing motion patterns by applying geometric transformations to isolated frames is significantly less effective, and that motion cues help much more when they are extracted from videos.
% In this work we aim at facilitating the training of highly-elaborate image interpretation algorithms, that are able to associate image pixels with surface coordinates.
% We explore different methods of improving supervision for dense human pose estimation tasks by leveraging on weakly-supervised and self-supervised learning. We  introduce DensePose-Track, a novel dataset for Dense Pose estimation in time, and exploit the interplay between image correspondence and surface analysis. 
% Exploiting temporal information allows us to improve upon strong baselines, delivering substantially improved dense pose estimation results. 
\end{abstract}\vspace*{-7mm}
\footnotetext[1]{James Thewlis and Iasonas Kokkinos were with Facebook AI Research (FAIR) during this work.}
\pagebreak

\vspace*{-3mm}
\section{Introduction}\label{s:intro}

The analysis of people in images and videos is often based on landmark detectors, which only provide a sparse description of the human body via keypoints such as the hands, shoulders and ankles.
More recently, however, several works have looked past such limitations, moving towards a combined understanding of object categories, fine-grained deformations~\cite{JaderbergSZK15,PapandreouKS15,DaiQXLZHW17,neverova} and  \emph{dense geometric structure}~\cite{guler2016densereg,ZhouKAHE16,denseiccv17,densepose,angjoo1,angjoo2,ThewlisBV17a}.
Such an understanding may arise via fitting complex 3D models to images or, as in the case of DensePose~\cite{densepose}, in a more data-driven manner, by mapping images of the object to a dense UV frame describing its surface.
%DensePose (DP)~\cite{densepose} is a prominent example of the latter approach, demonstrating the potential of modern deep networks to learn the dense intrinsic geometry of deformable object categories such as people.

%\input{fig-splash}

%share the same practical limitations of most modern machine learning approaches in image understanding, namely the 
Despite these successes, most of these techniques 
need large quantities of annotated data for training, proportional to the complexity of the model.
%In fact, as the sophistication of the model increases,  annotating data becomes more complex.
For example, in order to train DensePose, the authors introduced an intricate annotation framework and used it to crowd-source manual annotations for 50K people, marking a fairly dense set of landmark points on each person, for a grand total of 5M manually-labelled 2D points.
%The cost of the DensePose dataset is estimated to be 30K\,\$.
Such an extensive annotation effort is justified for visual objects such as people that are particularly important in numerous applications, but these methods cannot reasonably scale up to a dense understanding of the whole visual world.

Aiming at solving this problem, papers such as~\cite{ThewlisBV17a,dae} have proposed models similar to DensePose, but replacing manual annotations with self-supervision~\cite{ThewlisBV17a} or even no supervision~\cite{dae}.
The work of~\cite{ThewlisBV17a}, in particular, has demonstrated that a dense object frame mapping can be learned for simple objects such as human and pet faces using nothing more than the compatibility of the mapping with synthetic geometric transformations of the image, a property formalized as the \emph{equivariance} of the learned mapping.
Nevertheless, these approaches typically fail to learn complex articulated objects such as people.

In this paper, we thus examine the interplay between weakly-supervised and self-supervised learning with the learning of complex dense geometric models such as DensePose (\cref{f:splash}). Our goal is to identify a strategy that will allow us to use the least possible amount of supervision, so as to eventually scale models like DensePose to more non-rigid object categories. 
%how performance changes when using a considerably smaller quantity of manual annotations.

We start by exploring the use of sources of weaker supervision, such as semantic segmentation masks and human keypoints.
%Annotating selected video frames can be seen as a way to sparsify annotations through time.
%We also explore the complementary approach of sparsifying annotations through space.
In fact, one of the key reasons why collecting annotations in DensePose is so resource-intensive is the sheer amount of points that need to be manually clicked to label image pixels with surface coordinates. 
By contrast, masks and keypoints do not require establishing correspondences and as such are a lot easier to collect.
We show that, even though keypoints and masks alone are insufficient for establishing correct UV coordinate systems, they allow us to substantially sparsify the number of image-to-surface correspondences required to attain a given performance level.
%%%

We then extend the idea of sparsifying annotations to the temporal domain and turn to annotating selected video frames
in a video instead of still images as done by~\cite{densepose}. 
%can be seen as a way
%to sparsify annotations through time. We also explore
%the complementary approach of sparsifying annotations through space
%We then explore the benefit of annotating frames in a video .
For this we introduce DensePose-Track, a large-scale dataset consisting of dense image-to-surface correspondences gathered on the sequences of frames comprising the PoseTrack dataset~\cite{PoseTrack}.
While the process of manually annotating a video frame is no different than the process of annotating a similar still image, videos contain motion information that, as we demonstrate, can be used to multiply the efficacy of the annotations.
%%%
In order to do so, we use an off-the-shelf algorithm for optical flow~\cite{ilg2017flownet} to establish reliable \emph{dense correspondence} between different frames in a video.
We then use these correspondences in two ways: to \emph{transfer annotations} from one frame to another and to enforce an \emph{equivariance constraint} similar to~\cite{ThewlisBV17a}.

We compare this strategy to the approach adopted by several recent papers~\cite{ThewlisBV17a, zhang2018unsupervised, thewlis2017unsupervised} that use for this purpose synthesized image transformations, thus replacing the actual object deformation field with simple rotations, affine distortions, or thin-plate splines (TPS).
Crucially, we demonstrate that, while synthetic deformations are not particularly effective for learning a model as complex as DensePose, data-driven flows work well, yielding a strong improvement over the strongest existing baseline trained solely with manually collected static supervision.  
%%%
% In order to train and test our models we introduce DensePoseTrack, a  large-scale dataset consisting of dense image-surface correspondences gathered on the images of the PoseTrack dataset. We examine the impacts of different datasets on training performance, and propose a dataset mixing  approach to exploit from both static and spatio-temporal datasets.

\section{Related work}\label{s:realted}

Several recent works have aimed at reducing the need for strong supervision in fine-grained image understanding tasks. 
In semantic segmentation for instance \cite{gpapanweak,scribble,schielesimple} successfully used weakly- or semi- supervised learning in conjunction with low-level image segmentation techniques. Still, semantic segmentation falls short of delivering a surface-level interpretation of objects, but rather acts as a dense, `fully-convolutional' classification system. 

%Starting from the assumption 
Starting from a more geometric direction, several works have aimed at establishing dense correspondences between pairs \cite{BristowVL15} or sets of RGB images, as e.g. in the recent works of  \cite{ZhouKAHE16,denseiccv17}. More recently, \cite{ThewlisBV17a} use the equivariance principle in order to align sets of images to a common coordinate system, while \cite{dae} showed that autoencoders could be trained to reconstruct images in terms of templates deformed through UV maps. More recently, \cite{angjoo1} showed that silhouettes and landmarks suffice to recover 3D shape information when used to train a 3D deformable model. These approaches bring unsupervised, or self-supervised learning closer to the deformable template paradigm \cite{hands,cootes1998active,BlVe03}, that is at the heart of the connecting images with surface coordinates. Along similar lines, equivariance to translations was recently proposed in the context of sparse landmark localization in \cite{yaser}, where it was shown that it can stabilize network features and the resulting detectors.

%Our work draws inspiration

%These works however do not have the notion of
%a reference coordinate system (`template') to which images can get mapped - this makes the image generation and manipulation harder. , along the lines of the deformable template paradigm  

%Several works have aimed at incorporating deformations and alignment in a supervised setting, including Spatial Transformers  and DensePose~\cite{densepose}. These works have shown that  one can  improve accuracy of both classification and localization tasks by injecting deformations and  alignment within traditional CNN architectures. 
%works recover only a coarse-level interpretation of the 

%[COPY -PASTED FROM DAE PAPER - MIGHT GET BUSTED - NEED TO SMOOTH]

%Turning to unsupervised deep learning, even though most works focus on rigid, or low-dimensional parametric deformations, e.g. \cite{hinton10,WorrallGTB16},
%several works have attempted to incorporate richer non-rigid deformations within learning. A thread of works has been aimed at dynamically rerouting the processing of information within the network's graph based on the input, starting from neural computation arguments
%\cite{Hinton81,OlshausenAE95,vdm81}
%and eventually translating into concrete algorithms, such as the `capsule' works of \cite{Hinton11,Hinton17} that bind neurons on-the-fly. Still, these works lack a transparent, parametric handling of non-rigid deformations. 

\section{Method}\label{s:method}

We first summarise the DensePose model and then discuss two approaches significantly increasing the efficiency of collecting annotations for supervising this model.

\subsection{UV maps}\label{s:densepose}

DensePose can be described as a dense body landmark detector.
In landmark detection, one is interested in detecting a discrete set of body landmarks $u=1,\dots,U$, such as the shoulders, hands, and knees.
Thus, given an image $I:\mathbb{R}^2\rightarrow \mathbb{R}^3$ that contains a person (or several), the goal is to tell for each pixel $p\in\mathbb{R}^2$ whether it contains any of the~$U$ landmarks and, if so, which ones.

DensePose generalizes this concept by considering a dense space of landmarks $\mathcal{U} \subset \mathbb{R}^2$, often called a UV-space.
It then learns a function $\Phi$ (a neural network in practice) that takes an image $I$ as input and returns an association of each pixel $p$ to a UV point $u = \Phi_p(I) \in \mathcal{U} \cup \{\phi\}$.
Since some pixels may belong to the background region instead of a person, the function can also return the symbol $\phi$ to denote background.
The space $\mathcal{U}$ can be thought of as a ``chart'' of the human body surface; for example, a certain point $u\in \mathcal{U}$ in the chart may correspond to ``left eye'' and another to ``right shoulder''. In practice the body is divided into multiple charts, with a UV map predicted per part.

While DensePose is  more powerful than a traditional landmark detector, it is also  more expensive to train.
In traditional landmark detectors, the training data consists of a dataset of example images $I$ where landmarks are manually annotated;
the conceptually equivalent annotations for DensePose are UV associations $\Phi^\text{gt}_p(I) \in \mathcal{U}$ collected densely for every pixel $p$ in the image.
It is then possible to train the DensePose model $\Phi$ via minimization of a loss of the type $\|\Phi(I) - \Phi^{\text{gt}}(I)\|$.

In practice, it is only possible to manually annotate a discretized version of the UV maps.
Even so, this requires annotators to click hundreds of points per person instance, while facing issue with ambiguities in labeling pixels that are not localized on obvious human features (e.g.~points on the abdomen).
A key innovation of the DensePose work~\cite{densepose} was a new system to help human annotators to collect efficiently such data.
Despite these innovations, the DensePose-COCO training dataset consists of 50k people instances, for which 5 million points had to be clicked manually.
Needless to say, this required effort makes DensePose difficult to apply to new object categories.

\subsection{Geometric properties of the UV maps}\label{s:invariance}

Brute force manual labelling can be reduced by exploiting properties of the UV maps that we know must be satisfied a-priori.
Concretely, consider two images $I$ and $I'$ and assume that pixels $p$ and $p'$ in the respective images contain the same body point (e.g.~a left eye). 
Then, by definition, the map $\Phi$ must send pixels $p$ and $p'$ to the same UV point, so that we can write:
\begin{equation}\label{e:corresp}
  \Phi_p(I) = \Phi_{p'}(I').
\end{equation}
Consider now the special case where $I$ and $I'$ are frames of a video showing people deforming smoothly (where viewpoint changes are a special case of 3D deformation).
Then, barring self-occlusions and similar issues, corresponding pixels $(p,p')$ in the two images are related by a \emph{correspondence field} $g : \mathbb{R}^2 \rightarrow \mathbb{R}^2$ such that we can write $p' = g(p)$.
To a first approximation (i.e.~assuming Lambertian reflection and ignoring occlusions, cast shadows, and other complications) image $I'$ is a deformation $gI$ of image $I$ (i.e.~
$
  \forall p' :  (gI)(p') = I(g^{-1}(p'))
$%
). In this case, the compatibility equation~\eqref{e:corresp} can be rewritten as the so-called \emph{equivariance constraint}
\begin{equation}\label{e:equivariance}
\Phi_p(g I) = \Phi_{g(p)}(I)
\end{equation}
which says that the geometric transformation $g$ ``pops-out'' the function $\Phi$.

Next, we discuss how equivariance can be used in different ways to help supervise the DensePose model.
There are two choices here: (1) how the correspondence field $g$ can be obtained (\cref{s:g}) and (2) how it can be incorporated as a constraint in learning (\cref{s:loss}).

\begin{figure}[t!]
\includegraphics[width=0.9\linewidth]{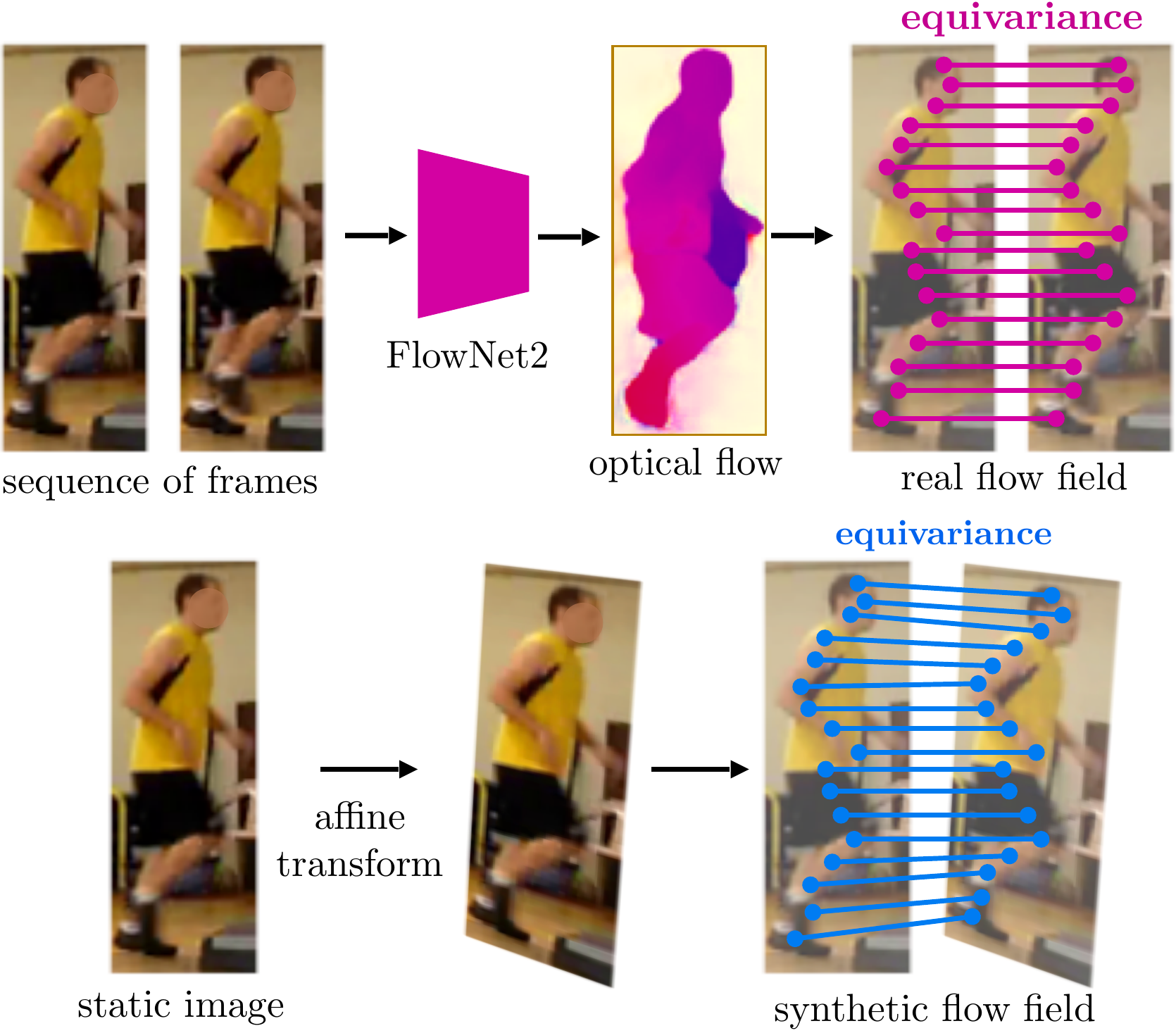}%\vspace*{5mm}
\\
\caption{Real (top) and synthetic (bottom) transformation fields exploited to enforce equivariance constraints.}\label{fig:flowprop}
\end{figure}
\subsubsection{Correspondence fields: synthesized vs real}\label{s:g}

%The invariance constraint~\eqref{e:equivariance} requires two images $I$ and $I'$ together with a correspondence field $g : \mathbb{R}^2 \rightarrow \mathbb{R}^2$ telling, for each pixel $p$ in the first image, what is the corresponding pixel $p'=gp$ in the second.
Annotating the correspondence field $g$ in~\eqref{e:equivariance} is no easier than collecting the DensePose annotations in the first place.
Thus, \eqref{e:equivariance} is only useful if correspondences can be obtained in a more efficient manner.
In this work, we contrast two approaches: synthesizing correspondences or measuring them from a video (see \cref{fig:flowprop}).

The first approach, adopted by a few recent papers~\cite{ThewlisBV17a, zhang2018unsupervised, thewlis2017unsupervised}, \emph{samples} $g$ at random from a distribution of image warps.
Typical transformations include affinities and thin-plate splines (TPS).
Given the warp $g$, a training triplet $t=(g, I, I')$ is then generated by taking a random input image $I$ and applying to it the warp to obtain $I'=gI$.

The second approach is to \emph{estimate} a correspondence field from data.
This can be greatly simplified if we are given a video sequence, as in this case low-level motion cues can be integrated over time to give us correspondences.
The easiest way to do so is to apply to the video an off-the-shelf optical flow method, possibly integrating its output over a short time. Then, a triplet is formed by taking the first frame $I$, the last $I'$, and the integrated flow $g$.

The synthetic approach is the simplest and most general as it does not require video data.
However, sampled transformations are at best a coarse approximation to correspondence fields that may occur in nature; in practice, as we show in the experiments, this severly limits their utility.
On the other hand, measuring motion fields is more complex and requires video data, but results in more realistic flows, which we show to be a key advantage.

\subsubsection{Leveraging motion cues}\label{s:loss}

Given a triplet $t=(g,I,I')$, we now discuss two different approaches to generating a training signal: transferring ground-truth annotations and Siamese learning.

The first approach assumes that the ground-truth UV map $\Phi_{p'}^\text{gt}(I')$ is known for image $I'$, as for the DensePose-Track dataset that will be introduced in~\cref{sec:dataset}.
Then, \cref{e:equivariance} can be used to recover the ground-truth mapping for the first frame $I$ as
$\Phi_p^\text{gt}(I) = \Phi_{g(p)}^\text{gt}(I')$.
In this manner, when training DP, the loss term $\|\Phi^\text{gt}(I') - \Phi(I')\|$ can be augmented with the term $\|\Phi^\text{gt}(I) - \Phi(I)\|$.

The main restriction of the approach above is that the ground-truth mapping must be available for one of the frames.
Otherwise, we can still use~\cref{e:equivariance} and enforce the constraint $\Phi_p(I) = \Phi_{gp}(I')$.
This can be encoded in a loss term of the type $\|\Phi(I) - \Phi_{g}(I'))\|$ where $\Phi_{g}(I')$ is the warped UV map of the second image.
Note that both terms in the loss are output by the learned model $\Phi$, which makes this a Siamese neural network configuration.

Another advantage of the equivariance constaint~\cref{e:equivariance} is that it can be applied to \emph{intermediate layers} of the deep convolutional neural network $\Phi$ as in fact the nature of the output of the function is irrelevant. % (\cref{e:equivariance} only cares about equality).
In the experiments, equivariance is applied to the features preceding the output layers at each hourglass stack as this was found to work best.
Thus, denote by $\Psi(I)$ the tensor output obtained at the appropriate layer of network $\Phi$ with input $I$ and let $\Psi_g$ be the warped tensor.
We encode the equivariance constraint via the cosine-similarity loss of the embedding tensors
$
\mathcal{L}_{\cos} = 1 - \rho(\Psi(I), \Psi_{g}(I')),
$
where~$\rho$ is the cosine similarity $\rho(x,y) = \langle x, y\rangle / (\|x\|\|y\||)$ of vectors $x$ and $y$. %\pagebreak

\section{DensePose-Track}\label{sec:dataset}

%Training our model requires supervision for  dense image-surface correspondence over time. This is clearly a challenging supervision signal, requiring extensive and laborious manual annotations. 

%Most existing annotation efforts for human pose in video focus on a sparse set of 10-20  body landmarks, such as wrists, elbows or knees.  Earlier video pose datasets, such as J-HMDB \cite{Jhuang:ICCV:2013}, Penn Action~\cite{zhang2013actemes} or more recent ones such as YouTube Pose~\cite{charles2016personalizing} often focus on single isolated individuals.  Similarly, the ``HumanEva'' \cite{Sigal:IJCV:10b} and ``Human3.6M'' \cite{ionescu2014human3} datasets provide 3D pose information of humans in  videos but are recorded in  controlled indoor environments.

We introduce the DensePose-Track dataset, based on the publicly available version of the PoseTrack dataset~\cite{PoseTrack}, which contains 10\,339 images and 76\,058 annotations.
PoseTrack annotations are provided densely for 30 frames that are located temporally in the middle of the video.
The DensePose-Track dataset has 250 training videos and 50 validation videos.
In order to allow a more diverse evaluation of long range articulations, every fourth frame is additionally annotated for the validation set.

%Since the data collection procedure for dense surface correspondences is much more complex and time consuming with respect to skeleton-based pose,
Since subsequent frames in DensePose-Track can be highly correlated, we temporally subsample the tracks provided in the PoseTrack dataset using two different sampling rates.
Firstly, in order to preserve the diversity and capture slower motions, we annotate every eighth frame.
Secondly, in order to capture faster motions we sample every second frame for four frames in each video.

Each person instance in the selected images is cropped based on a bounding box obtained from the keypoints and histogram-equalized.
The skeleton is superimposed on the cropped person images to guide the annotators and identify the person in occlusion cases. The collection of correspondences between the cropped images and the 3D model is done using the efficient annotation process analogous to the one described in~\cite{densepose}.

\begin{figure}
\begin{center}
\includegraphics[width=1.\linewidth]{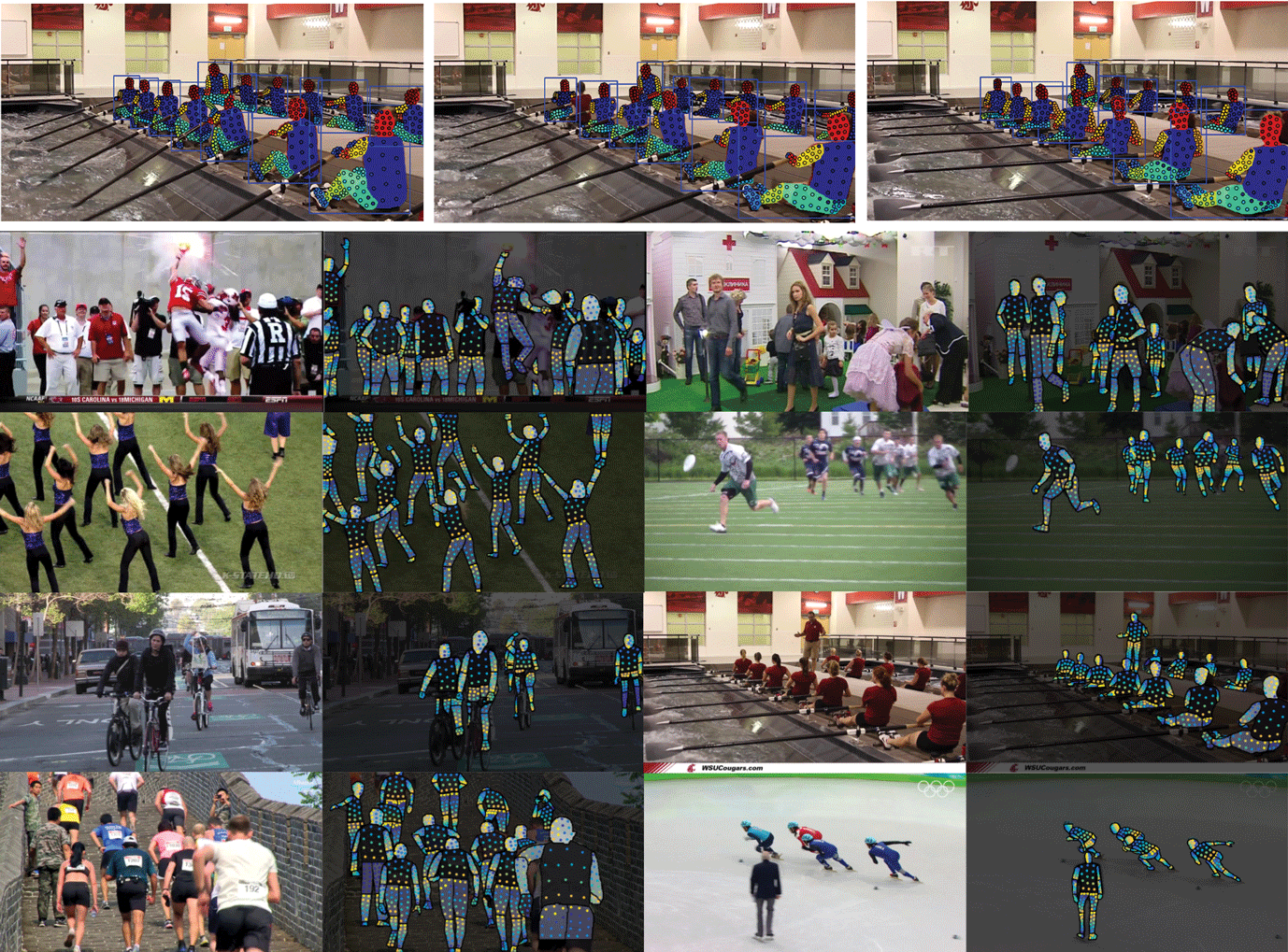}%\vspace*{1mm}
\caption{Annotations in the collected DensePose-Track dataset. 
Top row: Parts and points. 
Bottom rows: Images and collected points colored based on the `U' value~\cite{densepose}, indicating one of the two coordinates in a  part-based, locally planar parameterization of the human body surface.}\label{fig:dataset}
\end{center}
\end{figure}

The PoseTrack videos  contain rapid motions, person occlusions and scale variation which leads to a very challenging annotation task.
Especially due to motion blur and small object sizes, in many of the cropped images the visual cues are enough to localize the keypoints but not the detailed surface geometry.
To cope with this we have filtered the annotation candidates.
Firstly, the instances with less then six visible keypoints are filtered out. This is followed by manual elimination of samples that are visually hard to annotate.

The DensePose-Track dataset training/validation sets have 1680\,/\,782 images labelled in total with dense correspondences for 8274\,/\,4753 instances, resulting in a total of 800\,142\,/\,459\,348 point correspondences, respectively. 
Sample image-annotation pairs are visualized in Fig~\ref{fig:dataset}. %We note that the annotators are not provided by consecutive frames and do not need to track points - which would render the annotation substantially harder. Instead we randomly sample points

Beyond the purpose of self-supervised training through optical flow, PoseTrack contains information that could be used to asses dense pose estimation systems in time, or improve them through spatio-temporal processing at test time.
Static datasets do not capture the effects of occlusions caused by multi-person interactions, e.g.\ when dancing.
Recent datasets for  pose estimation in time focus on more challenging, multi-person videos as e.g.\ \cite{iqbal2017posetrack,insafutdinov2017arttrack}, but are smaller in scale --- in particular due to the challenging nature of the task.
Regarding establishing dense correspondences between images and surface-based body models DensePose-COCO was introduced in~\cite{densepose}, providing annotations for 50K images of humans appearing in the COCO dataset.
Still, this dataset only contains individual frames, and as such  cannot be used to train models that exploit  temporal information.
We intend to explore these research avenues in future work, and  focus here on studying how to best exploit temporal information as a means of supervision. 

\begin{table*}[!t]
\begin{center}
\footnotesize
\begin{tabular}{lccp{4mm}p{6mm}p{7mm}p{6mm}p{7mm}p{4mm}p{6mm}p{7mm}p{6mm}p{7mm}}%cc
\toprule
\hspace*{\fill} Model \hspace*{\fill}  & Train & Test & \emph{$\mathbf{AP}$} & \emph{$\mathbf{AP}_{50}$} & \emph{$\mathbf{AP}_{75}$} & \emph{$\mathbf{AP\!}_{M}$} & \emph{$\mathbf{AP}_{L}$}  & \emph{$\mathbf{AR}$} & \emph{$\mathbf{AR}_{50}$} & \emph{$\mathbf{AR}_{75}$} & \emph{$\mathbf{AR}_{M}$} & \emph{$\mathbf{AR}_{L}$} \\
\midrule        
%$\vphantom{\rule{0mm}{3.5mm}}$%
%\multicolumn{11}{c}{$\vphantom{\rule{0mm}{2mm}}$% 
%\emph{ResNet-18 backbone}}\\ \hline
%$\vphantom{\rule{0mm}{4mm}}$%
DensePose-RCNN  & DP-COCO & DP-COCO & 55.5 & 89.1 & 60.8 & 50.7 & 56.8 & 63.2 & 92.6 & 69.6 & 51.8 & 64.0 \\
Hourglass  & DP-COCO & DP-COCO & 57.3 & 88.4 & 63.9 & 57.6 & 58.2 & 65.8 & 92.6 & 73.0 & 59.6 & 66.2 \\
%Hourglass  & DP-COCO & DP-COCO &  &  &  &  &  &  &  &  &  &  \\
%DP-HG + GT prop. + equiv. &All & DP-COCO & 56.9 & 87.5 & 63.6 & 56.6 & 57.9 & 65.3 & 91.8 & 72.7 & 58.4 & 65.8 \\
\midrule%\midrule
DensePose-RCNN  & DP-COCO &DP-Track & 30.1 & 61.3 & 26.4 & 4.5 & 32.2 & 37.5 & 67.3 & 36.9 & 5.7 & 39.7\\
Hourglass  & DP-COCO & DP-Track & 39.3 & 70.7 & 38.9 & 22.4 & 40.6 & 48.7 & 78.3 &  50.8 & 33.2 & 49.8\\
\phantom{Hg} + GT prop. + equiv.& All & DP-Track & 40.3 & 72.3 & 39.7 & 23.3 & 41.6 & 49.4 & 79.5 & 51.6 & 34.1 & 50.5 \\\bottomrule
\end{tabular}\vspace*{-2mm}
\caption{Comparison with the state-of-the-art of dense pose estimation in a multi-person setting on DensePose-COCO (\texttt{DP-COCO}) and DensePose-Track (\texttt{DP-Track}) datasets. The DensePose-RCNN baseline is based on a ResNeXt-101 backbone, Hourglass has 6 stacks. In all cases we use real bounding box detections produced by DensePose-RCNN.} \label{tab:cocotable}
\end{center}
\end{table*}%\vspace{5mm}

\mycomment{
\begin{table}
\begin{center}
\footnotesize
\begin{tabular}{lcccc}
\toprule
\hspace*{\fill} Data \hspace*{\fill}& 5 cm & 10 cm & 20 cm \\
\midrule
Human (*) & 65 & 92 & 98 \\ 
%\midrule
DensePose-RCNN  & 51.16 & 68.21 & 78.37 \\ 
Hourglass -- 1 stack & \st{50.38} &  \st{77.97} & \st{89.80}  \\
\phantom{Hourglass --} 2 stacks & \st{55.78} &  \st{82.34} & \st{92.55}  \\
\phantom{Hourglass --} 8 stacks & \st{\textbf{58.23}} &  \st{\textbf{84.06}} & \st{\textbf{93.57}} \\
Hourglass -- 1 stack & 49.89 & 74.04 & 82.98 \\
\phantom{Hourglass --} 2 stacks & 52.23 & 76.50 & 84.99 \\
\phantom{Hourglass --} 8 stacks & 56.04 & 79.63 & 87.55 \\
\bottomrule
\end{tabular}
\end{center}
\caption{\textbf{Baseline architectures.}
Comparison of different DensePose architectures on the DensePose-COCO dataset (with ground truth bounding boxes): the original ResNeXt-based 
RCNN network of~\cite{densepose} and the Hourglass architecture~\cite{NewellYD16}.
Accuracy on the DensePose-COCO dataset increases with the number of hourglass stacks. However, deeper models overfit the biases of the COCO dataset used for pretraining, so that the best performance when transferred to DensePose-Track is at 6 stacks. (*) evaluated on manually annotated synthetic images\cite{densepose}.  
%[* with best performing ResNetXt-101 backbone]
}\label{tab:hourglass}
\end{table}\vspace*{1mm}
}
\begin{table}
\begin{center}
\footnotesize
\begin{tabular}{lccc|ccc}
\toprule
\multirow{2}{*}{\hspace*{\fill} Data \hspace*{\fill}} & \multicolumn{3}{c}{\hspace*{\fill} Full performance \hspace*{\fill}} & \multicolumn{3}{c}{\hspace*{\fill} Localization only \hspace*{\fill}}\\
& 5 cm\! & 10 cm\! & 20 cm & 5 cm\! & 10 cm\! & 20 cm \\
\midrule
Human (*) & 65 & 92 & 98 & -- & -- & --\\ 
%\midrule
DP-RCNN~\cite{densepose}\!\!\!\!  & 51.16\! & 68.21\! & 78.37\! & -- & -- & --\\ 
HG - 1 stack & 49.89\! & 74.04\! & 82.98\! & 50.38\! & 77.97\! & 89.80\\
\phantom{HG -} 2 stacks\!\! & 52.23\! & 76.50\! & 84.99 &55.78\! & 82.34\! &92.55 \\
\phantom{HG -} 8 stacks\!\! & 56.04\! & 79.63\! & 87.55 & 58.23\! & 84.06\! & 93.57\\
%
%Hourglass -- 1 stack & \st{50.38} &  \st{77.97} & \st{89.80}  \\
%\phantom{Hourglass --} 2 stacks & \st{55.78} &  \st{82.34} & \st{92.55}  \\
%\phantom{Hourglass --} 8 stacks & \st{\textbf{58.23}} &  \st{\textbf{84.06}} & %\st{\textbf{93.57}} \\
\bottomrule
\end{tabular}
\end{center}
\caption{\textbf{Baseline architectures.}
Comparison of different DensePose architectures on the DensePose-COCO dataset (given ground truth detections): the ResNeXt-based 
RCNN network of~\cite{densepose} and the Hourglass (HG) architecture~\cite{NewellYD16}.
%Accuracy on the DensePose-COCO dataset increases with the number of hourglass stacks. However, deeper models overfit the biases of the COCO dataset used for pretraining, so that the best performance when transferred to DensePose-Track is at 6 stacks. (*) evaluated on manually annotated synthetic images\cite{densepose}.  
%[* with best performing ResNetXt-101 backbone]
We report both full performance (with DP outputs masked with the foreground segmentation mask, as in~\cite{densepose}) and localization only performance (without masking). In the rest of the manuscript we report the "localization only" metric.
}\label{tab:hourglass}
\end{table}\vspace*{1mm}
\section{Experiments}\label{s:experiments}

In the first part of the experiments (\cref{s:exp.baseline}), we discuss the baseline DensePose architecture and find out a new ``gold-standard'' setup for this problem.

In the second part (\cref{s:exp.ablation}), we use the DensePose-COCO dataset to ablate the amount and type of supervision that is needed to learn dense pose estimation in static images.
In this manner, we clarify how much data annotations can be reduced without major changes to the approach.

Finally, in the last part (\cref{s:exp.track}) we explore the interplay with temporal information on the DensePose-Track dataset and study how optical flow can help increase the accuracy of dense pose estimation in ways which go beyond generic equivariance constraints.

\subsection{Baseline architectures}\label{s:exp.baseline}

In most of the following experiments we consider the performance of dense pose estimation obtained on ground-truth bounding boxes in a single person setting (including the DensePose-RCNN evaluation).
This allows us to isolate any issues related to object detection performance, and focus exclusively on the task of dense image-surface alignment.
We further introduce the use of Hourglass networks~\cite{NewellYD16} as a strong baseline that is trained from scratch on the task of dense pose estimation.
This removes any dependency on pretraining on ImageNet, and allows us to have an orderly ablation of our training choices.
In this setting, we evaluate the performance in terms of ratios of points localized within 5\,cm, 10\,cm and 20\,cm from the ground truth position measured along the surface of the underlying 3D model (geodesic distance)~\cite{densepose}.

%As in~\cite{densepose}, we train Hourglass networks on two tasks, semantic part segmentation, and chart coordinate regression within each part.
%We use the experimental same choices as in~\cite{densepose} apart from longer learning rates required to train Hourglass networks from scratch.\todo{For Natalia: please correct}
\begin{figure}[t]
\centering
\includegraphics[height=4.2cm]{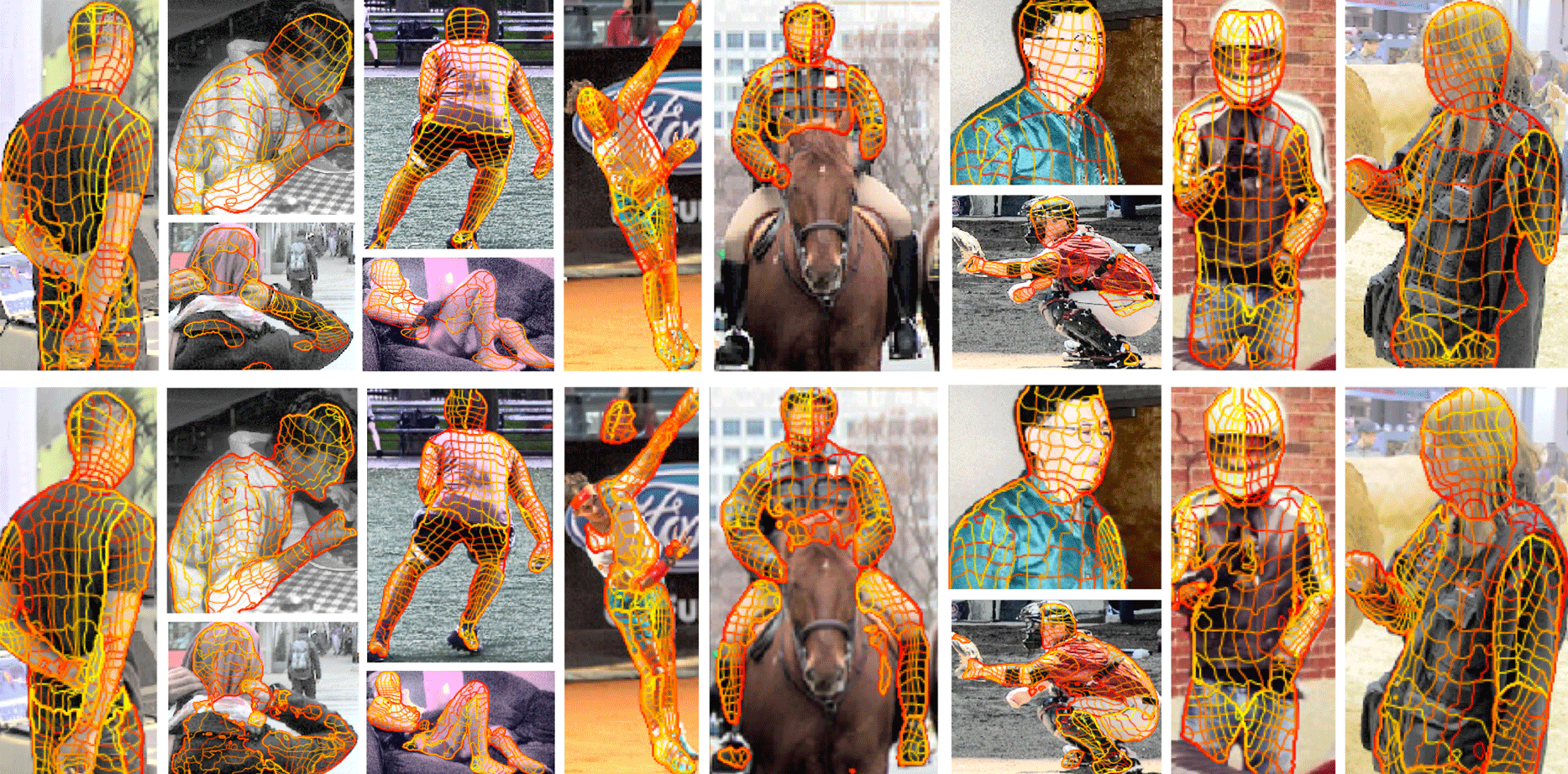}\vspace*{3mm}\\
\caption{\textbf{Qualitative results.} Hourglass (bottom) vs DensePose-RCNN~\cite{densepose} (top). The advantages of the fully convolutional Hourglass include better recall and spatial alignment of predictions with the input, at cost of higher sensitivity to high-frequency variations in textured inputs.}\label{fig:dpvshg}
\end{figure}%\vspace*{5mm}

Starting from the results in~\cref{tab:hourglass}, we observe that we get substantially better performance than the system of~\cite{densepose} which relies on the DensePose-RCNN architecture.
We note that the system of~\cite{densepose} was designed to execute both detection and dense pose estimation and operates at multiple frames per second; as such the numbers are not directly comparable. We do not perform detection, and instead report all results on images pre-cropped around the subject. Still, it is safe to conclude that Hourglass networks provide us with a strong baseline (see~\cref{fig:dpvshg} for illustrations).

For completeness, in~\cref{tab:cocotable} we also report performance of both architectures (DensePose-RCNN and Hourglass) in the multi-person setting, expressed in COCO metrics and obtained using the real bounding box detections produced by DensePose-RCNN with a ResNeXt-101 backbone.

%\Cref{fig:hgtransfer} further investigates the effect of the depth of Hourglass and its transferability from DP-COCO pretraining to DP-Track, showing that the best configuration is obtained at a depth of 6 blocks.

\subsection{Ablating annotations}\label{s:exp.ablation}

We first examine the impact of reducing the amount of DensePose supervision; we also consider using simpler annotations related to semantic part segmentation that are faster to collect than DensePose chart annotations.

\paragraph{Reduced supervision.}

Recall that DensePose annotations break down the chart $\mathcal{U}=\cup_{k=1}^K \mathcal{U}_k \subset \mathbb{R}^2$ into $K$ parts and, for each pixel $p$, provide the chart index $k(p)$ (segmentation masks) and the specific chart point $u(p) \in \mathcal{U}_{k(p)}$ within it ($u(p)$ is in practice normalized in the range $[0,1]^2$).
The neural network $\Phi_p(I) \approx (k(p),u(p))$ is tasked with estimating both quantities in parallel, optimizing a classification and a regression loss respectively.

\begin{figure*}[t!]
\centering
\includegraphics[width=0.98\linewidth]{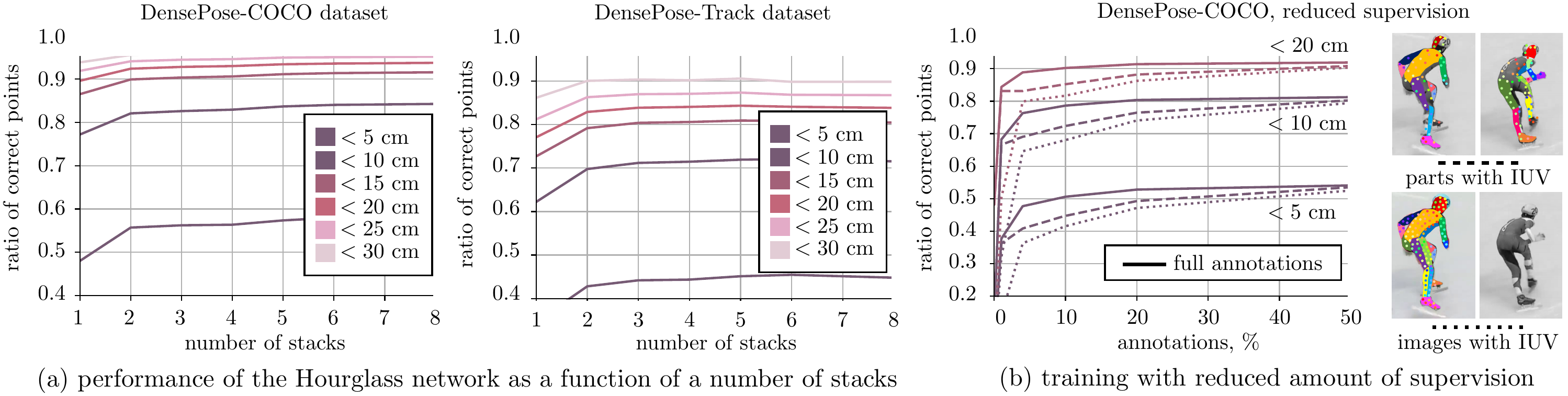}\vspace{1mm}%
\caption{(a) Performance of the Hourglass architecture on the DensePose-COCO dataset monotonically increases with the number of stacks, but peaks at 6 stacks for the DensePose-Track dataset. (b) Given a fixed annotation budget, it is beneficial to partially annotate a large number of images, rather than collect full annotations on a subset of the dataset.\label{fig:hgtransfer}}
\end{figure*}%\vspace*{5mm}

\begin{table}[t]
\footnotesize
\begin{center}
\begin{tabular}{clccc}
\toprule
&Data & 5 cm & 10 cm & 20 cm \\
\midrule
(i)&Full dataset & 55.78 & 82.34 & 92.55 \\ 
%(A01) Only masks & XX &  XX & XX & 3909078 \\ 
(ii)&Segmentation only & 3.53 & 13.25 & 48.21 \\
\midrule
(iii) %& \\ %\multicolumn{4}{c}{\textit{Subset of DP-COCO images}}\\
%\cmidrule(l){2-5}
& $50\%\,(k+u)$&  52.49 & 79.45 & 90.40 \\%& 0.57& 0.42 \\
%& 20\% images&  47.10 & 74.06 & 86.27 \\%& 0.57& 0.42 \\
%& 10\% images&  41.51 & 68.04 & 81.70 \\%& 0.57& 0.42 \\
image &  $5\%\,(k+u)$&  36.27 & 64.58 & 79.93 \\%& 0.57& 0.42 \\
subsampling&  $1\%\,(k+u)$&  14.11 & 32.06 & 50.21 \\%& 0.57& 0.42 \\
\midrule
(iv) %& \\ %\multicolumn{4}{c}{\textit{Subset of DP-COCO images with full supervision}}\\
%\cmidrule(l){2-5}
& $100\%\,k + 50\%\,u$&  53.50 & 80.29 & 90.86 \\%& 0.57& 0.42 \\
%&  20\% images with $u$&  49.26 & 76.47 & 88.21  \\%& 0.57& 0.42 \\
%&  10\% images with $u$&  44.71 & 72.38 & 85.22 \\%& 0.57& 0.42 \\
image & $100\%\,k + 5\%\,u$&  40.80 & 69.04 & 83.15 \\%& 0.57& 0.42 \\
subsampling& $100\%\,k + 1\%\,u$&  36.16 & 66.59 & 83.14\\%& 0.57& 0.42 \\
%\hline\hline
\midrule
(v) %& \\ %\multicolumn{4}{c}{\textit{Subset of DP-COCO points}}\\
%\cmidrule(l){2-5}
&  $50\%\,(k + u)$&  54.06 & 81.24 & 91.92 \\%& 0.57& 0.42 \\
%&  20\% $u$&  52.79 & 80.39 & 91.42 \\%& 0.57& 0.42 \\
%&  10\% $u$&  50.58 & 78.65 & 90.23 \\%& 0.57& 0.42 \\
point&  $5\%\,(k+u)$&  47.68 & 76.34 & 88.86 \\%& 0.57& 0.42 \\
subsampling &  $1\%\,(k+u)$&  37.65 & 68.25 & 84.37 \\%& 0.57& 0.42 \\
\bottomrule
\end{tabular}
\end{center}
\caption{Reduced supervision on DensePose-COCO, $k$ stands for body part index and $u$ for UV coordinates (\cref{fig:hgtransfer}b additionally illustrates experiments (i), (iii) and (v)).}\label{weakify}
\end{table}
%\begin{figure}[t]
%\includegraphics[width=\linewidth]{figures/keypoints.pdf}
%\end{figure}

\begin{table}[t]
\begin{center}
\footnotesize
\begin{tabular}{llccc}%c|c|}
\toprule
\multirow{5}[6]{*}{\includegraphics[width=1.15cm]{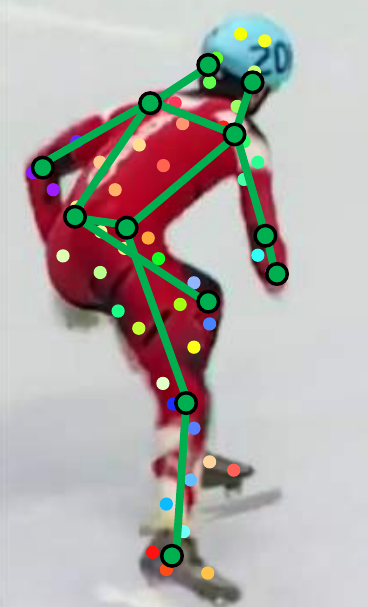}} &
Data & 5 cm & 10 cm & 20 cm \\
\cmidrule(l){2-5}
& Full dataset & 55.78 & 82.34 & 92.55 \\
\cmidrule(l){2-5}
%   100\% + kp & 44.79 &  XX & XX \\
 & 1\% $u$ &  37.65 & 68.25 & 84.37 \\%& 0.57& 0.42 \\
 & keypoints &  36.60 & 63.03 & 76.81 \\
 & 1\% $u$ + keypoints &  39.17 & 68.78 & 85.12 \\[1mm]%& 0.57& 0.42 \\
%(B03) all masks + kp + warpfeat &  XX & XX & XX & 3908949, 3909187, 3918125\\%& 0.57& 0.42 \\
\bottomrule
%\cmidrule(l){2-5}
\end{tabular}
\end{center}
\caption{The positive effect of augmenting sparse DensePose-COCO annotations with skeleton keypoints. }\label{kpt}
\end{table}

We first observe (rows (i) vs (ii) of~\cref{weakify}) 
that supervising only by using the segmentation masks (thus discarding the regression term in the loss) is not very useful, which is not surprising since they do not carry any surface-related information.
However, part masks can result in a much more graceful drop in performance when removing DensePose supervision.
To show this, in experiment (iii) we use only a subset of DensePose-COCO images for supervision (using complete part-point annotations $(k,u)$), whereas in (iv) we add back the other images, but only providing the simpler part $k$ annotations for the images we add back.
We see that performance degrades much more slowly, suggesting that, given an annotation budget, it is preferable to collect coarse annotations for a large number of images while collecting detailed annotations for a smaller subset.

The final experiment (v) in~\cref{weakify,fig:hgtransfer}b is similar, but instead of reducing the number of images, we reduce the number of pixels $p$ for which we provide chart point supervision $u(p)$ (thus saving a corresponding number of annotator ``clicks'').
For a comparable reduction in annotations, this yields higher accuracy as the network is exposed to a broader variety of poses during training.
Hence, for a fixed budget of annotator ``clicks'' one should collect fewer correspondences per image for a large number of images.

\begin{figure}
\centering
\includegraphics[height=4.4cm]{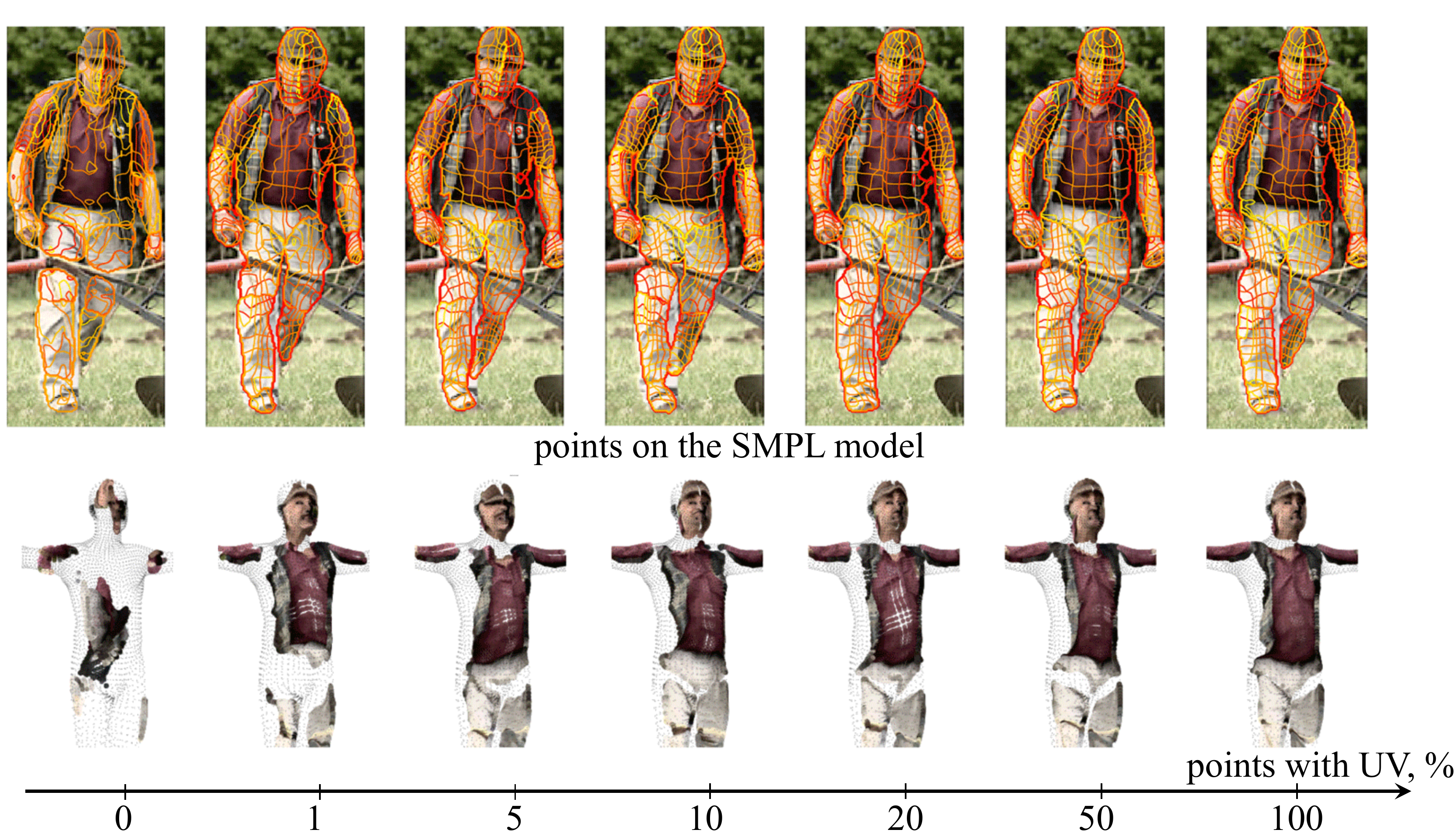}\vspace*{1mm}
\caption{\textbf{Reduced supervision.} Top: effect of training with a reduced percentage of points with UV annotations. Bottom: the texture map displayed on the SMPL model~\cite{bogo2016keep} shows the quality of the learned mapping.}\label{fig:samplingvisual}
\end{figure}

\paragraph{Keypoint supervision.}

Traditional landmark detectors use keypoint annotations only, which is even easier than collecting part segmentations.
Thus, we examine whether keypoint annotations are complementary to part segmentations as a form of coarse supervision.
In fact, since a keypoint associates a small set of surface points with a single pixel, this type of supervision could drive a more accurate image-surface alignment result. 
Note that not only keypoints are sparse, but they are also easier to collect from an annotator than an image-to-surface correspondence $u$, since they do not not require presenting to the annotator a clickable surface interface as done in~\cite{densepose}.
%; all that is required is clicking on an image. In particular every keypoint corresponds to an internal, 3D, skeleton joint and is associated with the subset of  surface nodes/triangles through which it can project to the image plane. This  defines the  permitted UV coordinates during training, which amounts to a form of multiple-instance-learning, with every keypoint being associated with a bag of UV values. 
%, while any other value is penalized as being wrong. 

%\todo{Explain how this is used in the loss}

\Cref{kpt} replicates the experiment (v.a) of~\cref{weakify}, repeats it but this time providing only keypoint annotations instead of $u$ annotations, and then combines the two.
We see that the two annotations types are indeed complementary, especially for highly-accurate localization regimes. %(localization within a 5 cm range).

\subsection{Paced learning}\label{s:exp.track}

Next, we examine statistical differences between the DensePose-COCO and DensePose-Track datasets (discarding for now dynamics) and their effect on training DensePose architectures.
We show that DensePose-Track does improve performance when used in combination with DensePose-COCO; however, it is substantially harder and thus must be learned in a paced manner, after the DensePose model has been initialized on the easier COCO data.

The details on this group of experiments are given in~\cref{tab:flow}. In all cases, we train a 6-stack Hourglass model, using the best performing architecture identified in the previous section. Stage~I means that the model is first initialized by training on the stated dataset and Stage~II, where applicable, means that the model is fine-tuned on the indicated data.
We observe that training on DensePose-Track (row (i) of \cref{tab:flow}) yields worse performance than training on an equiparable subset or the full DensePose-COCO dataset (ii-iii), even when the model is evaluated on DensePose-Track.
We assume that this is due to both the larger variability of images in the COCO training set, % (PoseTrack has many similar frames), 
as well as the cleaner nature of COCO images (blur-free, larger resolution),  which is known to assist training~\cite{BengioLCW09}.
This assumption is further supported by row (iv), where it is shown than training simultaneously on the union of COCO and PoseTrack yields worse results than training exclusively on COCO.

By contrast, we observe that a two-stage procedure, where we first train on DensePose-COCO and then finetune on DensePose-Track yields substantial improvements.
The best results are obtained by fine-tuning on the union of both datasets -- even giving an improvement on the DensePose-COCO test set.
This is again aligned with curriculum learning~\cite{BengioLCW09}, which suggests first training  on  easy examples and including  harder examples in a second stage.

\subsection{Leveraging motion cues}\label{s:motion}

Having established  a series of increasingly strong baselines, we now turn to validating the contribution of flow-based training when combined with the strongest baseline. % developed so far.

\paragraph{Flow computation.}

For optical flow computation we use the competitive neural network based method of FlowNet2~\cite{ilg2017flownet}, which has been trained on synthetic sequences.
We run this model on Posetrack and MPII Pose (video version), computing for each frame $T$ the flow to frames $T-3$ to $T+3$ (where available).
For MPII Pose we start with about a million frames and obtain 5.8M flow fields.
For DensePose-Track we have 68k frames and 390k flow fields.
Note that a subset of MPII Pose clips are used in DensePose-Track, although the Posetrack versions contain more frames of context.
For DensePose-Track, we propagate the existing DensePose annotations according to the flow fields, leading to 48K new cropped training images from the original 8K %(note that about 
(12\% of frames have manual labels).

In order to propagate annotations across frames, we simply translate the annotation locations according to the flow field~(\cref{fig:flowprop}).
Because optical flow can be noisy, especially in regions of occlusion, we use a forward-backward consistency check. If translating forward by the forward flow then back again using the backward flow gives an offset greater than 5 pixels we ignore that annotation.
On MPII pose, we use the annotations of rough person centre and scale. %, along with the ``single person'' flag. %, to obtain crops containing mainly isolated individuals.

\begin{table*}[!htb]
\begin{center}
\footnotesize
\begin{tabular}{clccccccc}
\toprule
&\multicolumn{2}{c}{Training data}&
\multicolumn{3}{c}{\!\!Tested on DensePose-Track}&
\multicolumn{3}{c}{\!\!\!Tested on DensePose-COCO}\\
\cmidrule(l){2-3}
\cmidrule(l){4-6}
\cmidrule(l){7-9}
&\hspace*{\fill}\emph{Stage I}\hspace*{\fill} & \hspace*{\fill}\emph{Stage II}\hspace*{\fill} & 5 cm & 10 cm & 20 cm & 5 cm & 10 cm & 20 cm\\
%\cmidrule(l){2-2}
%\cmidrule(l){3-3}
%\cmidrule(l){4-6}
%\cmidrule(l){7-9}
\midrule
%\multicolumn{4}{|c|}{$\vphantom{\rule{0mm}{3.5mm}}$ 
%\textit{Init COCO, Test-PoseTrack, 2stacks}}\\ \hline
%no train & 42.55 & 69.50 & 82.74  \\%& 0.57& 0.42 \\
%train-(PoseTrack) & 45.57  & 73.35 & 85.77 \\%& 0.57& 0.42 \\
% train-(PoseTrack+flowGT) & 45.98  & 73.57 & 86.11  \\%& 0.57& 0.42 \\
% train-(COCO+PoseTrack) &  43.87 & 70.59 & 83.81  \\%& 0.57& 0.42 \\
% train-(PoseTrack+flowGT+COCO) & 46.03  & 73.31 & 85.76  \\%& 0.57& 0.42 %\\
%\hline\hline
%\multicolumn{4}{|c|}{$\vphantom{\rule{0mm}{3.5mm}}$ 
%\textit{Init COCO, Test-PoseTrack, 6stacks}}\\ \hline
(i)&DensePose-Track & --- & 21.06 & 42.94 & 59.54 & 20.34 & 41.24 & 57.29  \\%& 0.57& 0.42 \
(ii)&DensePose-COCO subset (*) & --- & 33.67 & 58.79 & 73.45 & 47.10 & 74.06 & 86.27 \\%& 0.57& 0.42 \\
(iii)&DensePose-COCO & --- & 44.89 & 71.52 & 83.71 & 55.78 & 82.34 & 92.55 \\%& 0.57& 0.42 \\
(iv)&DensePose-COCO \& Track & --- & 41.76 & 69.94 & 83.60 & 55.27 & 82.05 & 92.37\\%& 0.57& 0.42 \\
\cmidrule(l){2-2}
\cmidrule(l){3-3}
\cmidrule(l){4-6}
\cmidrule(l){7-9}
(v)&DensePose-COCO & DensePose-Track & 45.57  & 73.35 & 85.77 & 53.70 & 81.34 & 92.03\\%& 0.57& 0.42 \\
%& train-(PoseTrack+wfGT) & 45.76  & 73.52 & 85.89 \\%& 0.57& 0.42 \\
%& train-(PoseTrack+flowGT) & 46.55  & 74.47 & 86.92 \\%& 0.57& 0.42 \\
(vi)&DensePose-COCO & all &  \textbf{46.04} & \textbf{73.41} & \textbf{85.79} & \textbf{58.01} & \textbf{84.06} & \textbf{93.64}\\%& 0.57& 0.42 \\
%DP-COCO & train-(PoseTrack+flowGT+COCO) & 47.28  & 74.62 & 86.72  \\%& 0.57& 0.42 \\
\bottomrule
\end{tabular}
\end{center}\vspace*{-2mm}
\caption{\textbf{Training strategies.} Effect of training on DensePose-COCO vs DensePose-Track in various combinations. The best performing model (vi) is first trained on the cleaner COCO data and then fine tuned on a union of datasets. (*) a random subset of the DensePose-COCO training images of size of DensePose-Track. (i)-(iv) have 2 stacks and (v)-(vi) have 8 stacks.}\label{tab:flow}
\end{table*}
\begin{table*}[!t]
\centering
\footnotesize
\begin{tabular}{clcccccccc}
\toprule
&
\multirow{2}{*}{\hspace*{8mm}Training strategy} &
\multicolumn{2}{c}{Training data}  &
\multicolumn{3}{c}{Synthetic\ (TPS)} &
\multicolumn{3}{c}{Real (Optical Flow)}
\\
\cmidrule(l){3-4}
\cmidrule(l){5-7}
\cmidrule(l){8-10}
& & 
\emph{Stage I} & \emph{Stage II} &
5 cm &
10 cm &
20 cm &
5 cm &
10 cm &
20 cm \\
\midrule
%\cmidrule(l){2-2}
%\cmidrule(l){3-3}
%\cmidrule(l){4-4}
%\cmidrule(l){5-7}
%\cmidrule(l){8-10}
(i)&Baseline & \multirow{4}{*}{---} & \multirow{4}{*}{\!\!\!DensePose-Track\!\!\!} & 21.06 & 42.94 & 59.54 & 21.06 & 42.94 & 59.54 \\
(ii)&GT propagation &&& 22.33 &  45.30 & 62.08 & 32.85 & 60.07 & 75.95\\
(iii)&Equivariance &&& 21.57 & 44.17 & 61.27 & 23.12 &  45.85 & 62.22\\
(iv)&GT prop. + equivariance\!\! &&&  \textbf{22.41} & \textbf{45.53} & \textbf{62.71} & \textbf{34.50} & \textbf{61.70} & \textbf{77.20}\\
\cmidrule(l){2-2}
\cmidrule(l){3-3}
\cmidrule(l){4-4}
\cmidrule(l){5-7}
\cmidrule(l){8-10}
(v)&Baseline & \multirow{4}{*}{\!\!DensePose-COCO\!\!} & \multirow{4}{*}{\!\!\!DensePose-Track\!\!\!} & 45.57  & 73.35 & 83.71 & 45.57  & 73.35 & 83.71 \\
(vi)&GT propagation &&& 45.77 & 73.65 & 86.13 & 47.36 & 75.17 & 87.47  \\
(vii)&  Equivariance &&& 45.67 & 73.47 & 85.93 & 45.76 & 73.54 & 86.06\\
(viii)&GT prop. + equivariance\!\! &&& \textbf{45.81} & \textbf{73.70} & \textbf{86.14} & \textbf{47.45} & \textbf{75.21} & \textbf{87.56} \\
\cmidrule(l){2-2}
\cmidrule(l){3-3}
\cmidrule(l){4-4}
\cmidrule(l){5-7}
\cmidrule(l){8-10}
(ix)&Baseline & \multirow{2}{*}{\!\!DensePose-COCO\!\!} & \multirow{2}{*}{all} & 46.04  & 73.41 & 85.79 & 46.04  & 73.41 & 85.79 \\
(x)&GT prop. + equivariance\!\! &&& \textbf{-} & \textbf{-} & \textbf{-} & \textbf{47.62} & \textbf{75.80} & \textbf{88.12}\\
\bottomrule
\end{tabular}\vspace*{1mm}\\
\caption{\textbf{Leveraging real and synthetic flow fields.} The best performing model (x) is trained on a combination  DensePose-COCO+Track by exploiting the real flow for GT propagation between frames and enforcing equivariance. }\label{tab:flow-augmented}
\end{table*}\vspace*{0.5mm}

%\cmidrule(l){2-2}
%\cmidrule(l){3-3}
%\cmidrule(l){4-6}
%\cmidrule(l){7-9}

\paragraph{Results.}

We compare the baseline results obtained in the previous section to different ways of augmenting training using motion information.
There are two axes of variations: whether motion is randomly synthesized or measured from a video using optical flow (\cref{s:g}) and whether motion is incorporated in training by propagating ground-truth labels or via the equivariance constraint (\cref{s:loss}).

%Results are presented in~\cref{tab:flow-augmented}.
Rows (i-iv) of~\cref{tab:flow-augmented} compare using the baseline supervision via the available annotations in DensePose-Track to their augmentation using GT propagation, equivariances and the combination of the two.
For each combination, the table also reports results using both synthetic (random TPS) and real (optical flow) motion.
Rows (v-viii) repeat the experiments, but this time starting from a network pre-trained on DensePose-COCO instead of a randomly initialized one.

There are several important observations.
First, both GT propagation and equivariance improve the results, and the best result is obtained via their combination.
GT propagation performs at least a little better than equivariance (but it cannot be used if no annotations are available).

Second, augmenting via real motion fields (optical flow) works a lot better than using synthetic transformations, suggesting that realism of motion augmentation is key to learn complex articulated objects such as people.

Third, the benefits of motion augmentation are particularly significant when one starts from a randomly-initialized network. If the network is pre-trained on DensePose-COCO, the benefits are still non-negligible. %,% but they are naturally more modest. % since the model is good to start with.

%\paragraph{Discussion.}
It may seem odd that GT propagation works better than equivariance since both are capturing similar constraints.
After analyzing the data, we found out that the reason is that equivariance optimized for \emph{some} charting of the human body, but that, since many charts are possible, this needs not to be the same that is constructed by the annotators.
Bridging this gap between manual and unsupervised annotation statistics is an interesting problem that is likely to be of relevance whenever such techniques are combined.

\paragraph{Equivariance at different feature levels.}

Finally, we analyze the effect of applying equivariance losses to different layers of the network, using synthetic or optical flow based transformations (see \cref{tab:equivlevel}). 
The results show benefits of imposing these constraints on the intermediate feature levels in the network, as well as on the subset of the output scores representing per-class probabilities in body parsing. % task. % of DensePose. 

\begin{table}[!t]
\begin{center}
\footnotesize
\begin{tabular}{lcccccc}
\toprule
\multirow{2}{*}{Features}&
\multicolumn{3}{c}{Synthetic (TPS)}& \multicolumn{3}{c}{Real (Optical Flow)} \\
\cmidrule(l){2-4}
\cmidrule(l){5-7}
 & 5 cm & 10 cm & 20 cm  & 5 cm & 10 cm & 20 cm \\
\midrule
0 & 45.74 & 73.62 & 86.14 & 45.90 & 73.71 & 86.10 \\ 
1 & \textbf{46.08} & \textbf{73.85} & \textbf{86.29} &45.91 & 73.74 & 86.15 \\ 
2 & 45.97 & 73.82 & \textbf{86.29} &45.92 & 73.64 & 86.04 \\ 
3 & 45.85 & 73.55 & 86.05 & 45.97 & 73.81 & \textbf{86.30} \\ 
4, all & 45.98 & 73.62 & 86.15 & 45.84 & 73.42 & 85.86 \\ 
4, segm. & 46.02 & 73.74 & 86.20 & \textbf{45.98} & \textbf{73.85} & 86.20\\
4, UV & 45.78 & 73.76 & 86.26 & 45.95 & 73.64 & 86.16 \\
none & 45.57 & 73.35 & 83.71 & 45.57 & 73.35 & 83.71 \\
% 00_FG0*_flowthresh_sgd_*_1.00_posetrackall_log_test.txt
\bottomrule
\end{tabular}
\end{center}\vspace*{-2mm}
\caption{Training with applying synthetic and optical flow warp-based equivariance at different feature levels (pre-training on DensePose-COCO, tuning and testing on DensePose-Track). Level 4 corresponds to the output of each stack, level 0 -- to the first layer. 'Segm.' denotes the segmentation part of the output, 'UV' -- the UV coordinates. }\label{tab:equivlevel}
\end{table}
% Recall from~\cref{s:} improvements: training using ground-truth propagation, training using equivariance
% We now compare the improvements we get by exploiting deformations during training. We use either synthetic deformations, obtained from Thin Plane Splines (TPS), or deformations established from optical flow-based correspondences (Flow). The deformations are used in complementary manners: (a) GT  propagation transfers ground truth from the annotated frames to the remaining ones (either synthetic for TPS or observed for Flow).
% (b) Equivariance enforces a dense correspondence-based equivariance constraint on the features of the network across the different frames.
% (c) GT+ Equivariance applies a combination.

% Our observations are consistent across datasets. Firstly, flow-based deformations clearly give larger improvements than synthetic, TPS-based deformations. This indicates the merit of using video for simplifying the training of dense pose estimation.

% Secondly, Ground-Truth propagation  yields the cleanest improvement - up to 11 points for Flow-PoseTrack. Thirdly, equivariance yields a small improvement  for Pose-Track, \todo{XXXX NOT CLEAR WHAT HAPPENS IN COCO WITH OPTICAL FLOW. NOT CLEAR WHAT HAPPENS WHEN COMBINIG GT + Equiv. XXXX}

\section{Conclusion}\label{s:conclusions}
In this work we have explored different methods of improving supervision for dense human pose estimation tasks by leveraging on weakly-supervised and self-supervised learning. This has allowed us to exploit temporal information to improve upon strong baselines, delivering substantially more advanced dense pose estimation results when compared to \cite{densepose}. We have also introduced a novel dataset DensePose-Track, which can facilitate further research at the interface of dense correspondence  and time. On the application side, this is crucial, for example, for enabling better user experiences in augmented or virtual reality.
%On the application side, this should lead to improved experiences in AR and gaming.
\par

Despite this progress, applying such models to videos on a frame-by-frame basis can reveal some of their shortcomings, including %potential redundancy in computations, 
flickering, missing body parts or false detections over the background (as witnessed in the hardest of the supplemental material videos). These problems can potentially be overcome by  exploiting temporal information, along the lines pursued in the pose tracking problem, \cite{Black00,BreglerMP04,insafutdinov2017arttrack,PoseTrack,3DCNN}, %For instance, motion blur, or partial occlusion can result in erroneous correspondences at a given image position; however, we can recover from such failures, 
i.e. by  combining complementary information from adjacent frames where the same structure is better visible. 
We intend to further investigate this direction in our future research. 

{\small\bibliographystyle{ieee_fullname}\bibliography{refs}}

\begin{thebibliography}{10}\itemsep=-1pt

\bibitem{BengioLCW09}
Yoshua Bengio, J{\'{e}}r{\^{o}}me Louradour, Ronan Collobert, and Jason Weston.
\newblock Curriculum learning.
\newblock In {\em ICML}, 2009.

\bibitem{BlVe03}
Volker Blanz and Thomas Vetter.
\newblock Face recognition based on fitting a 3{D} morphable model.
\newblock {\em PAMI}, 25(9):1063--1074, 2003.

\bibitem{bogo2016keep}
Federica Bogo, Angjoo Kanazawa, Christoph Lassner, Peter Gehler, Javier Romero,
  and Michael Black.
\newblock Keep it smpl: Automatic estimation of 3d human pose and shape from a
  single image.
\newblock In {\em ECCV}, 2016.

\bibitem{BreglerMP04}
Christoph Bregler, Jitendra Malik, and Katherine Pullen.
\newblock Twist based acquisition and tracking of animal and human kinematics.
\newblock {\em International Journal of Computer Vision}, 56(3):179--194, 2004.

\bibitem{BristowVL15}
Hilton Bristow, Jack Valmadre, and Simon Lucey.
\newblock Dense semantic correspondence where every pixel is a classifier.
\newblock In {\em ICCV}, 2015.

\bibitem{cootes1998active}
Timothy Cootes, Gareth Edwards, and Christopher Taylor.
\newblock Active appearance models.
\newblock In {\em ECCV}, 1998.

\bibitem{DaiQXLZHW17}
Jifeng Dai, Haozhi Qi, Yuwen Xiong, Yi Li, Guodong Zhang, Han Hu, and Yichen
  Wei.
\newblock Deformable convolutional networks.
\newblock In {\em ICCV}, 2017.

\bibitem{yaser}
Xuanyi Dong, Shoou{-}I Yu, Xinshuo Weng, Shih{-}En Wei, Yi Yang, and Yaser
  Sheikh.
\newblock Supervision-by-registration: An unsupervised approach to improve the
  precision of facial landmark detectors.
\newblock In {\em CVPR}, 2018.

\bibitem{denseiccv17}
Utkarsh Gaur and B.~S. Manjunath.
\newblock Weakly supervised manifold learning for dense semantic object
  correspondence.
\newblock In {\em ICCV}, 2017.

\bibitem{3DCNN}
Rohit Girdhar, Georgia Gkioxari, Lorenzo Torresani, Manohar Paluri, and Du
  Tran.
\newblock Detect-and-track: Efficient pose estimation in videos.
\newblock In {\em CVPR}, 2018.

\bibitem{hands}
U. Grenander, Y. Chow, and D.~M. Keenan.
\newblock {\em Hands: A Pattern Theoretic Study of Biological Shapes}.
\newblock Springer-Verlag, Berlin, Heidelberg, 1991.

\bibitem{densepose}
Riza~Alp G{\"{u}}ler, Natalia Neverova, and Iasonas Kokkinos.
\newblock Densepose: Dense human pose estimation in the wild.
\newblock In {\em CVPR}, 2018.

\bibitem{guler2016densereg}
R{\i}za~Alp G{\"u}ler, George Trigeorgis, Epameinondas Antonakos, Patrick
  Snape, Stefanos Zafeiriou, and Iasonas Kokkinos.
\newblock Densereg: Fully convolutional dense shape regression in-the-wild.
\newblock In {\em CVPR}, 2017.

\bibitem{ilg2017flownet}
Eddy Ilg, Nikolaus Mayer, Tonmoy Saikia, Margret Keuper, Alexey Dosovitskiy,
  and Thomas Brox.
\newblock Flownet 2.0: Evolution of optical flow estimation with deep networks.
\newblock In {\em CVPR}, 2017.

\bibitem{insafutdinov2017arttrack}
Eldar Insafutdinov, Mykhaylo Andriluka, Leonid Pishchulin, Siyu Tang, Evgeny
  Levinkov, Bjoern Andres, and Bernt Schiele.
\newblock Arttrack: Articulated multi-person tracking in the wild.
\newblock In {\em CVPR}, 2017.

\bibitem{PoseTrack}
Umar Iqbal, Anton Milan, Mykhaylo Andriluka, Eldar Ensafutdinov, Leonid
  Pishchulin, Juergen Gall, and Bernt Schiele.
\newblock Pose{T}rack: {A} benchmark for human pose estimation and tracking.
\newblock In {\em CVPR}, 2018.

\bibitem{iqbal2017posetrack}
Umar Iqbal, Anton Milan, and Juergen Gall.
\newblock Posetrack: Joint multi-person pose estimation and tracking.
\newblock In {\em CVPR}, 2017.

\bibitem{JaderbergSZK15}
Max Jaderberg, Karen Simonyan, Andrew Zisserman, and Koray Kavukcuoglu.
\newblock Spatial transformer networks.
\newblock In {\em NIPS}, 2015.

\bibitem{angjoo2}
Angjoo Kanazawa, Michael~J. Black, David~W. Jacobs, and Jitendra Malik.
\newblock End-to-end recovery of human shape and pose.
\newblock In {\em CVPR}, 2018.

\bibitem{angjoo1}
Angjoo Kanazawa, Shubham Tulsiani, Alexei~A. Efros, and Jitendra Malik.
\newblock Learning category-specific mesh reconstruction from image
  collections.
\newblock In {\em ECCV}, 2018.

\bibitem{schielesimple}
Anna Khoreva, Rodrigo Benenson, Jan~Hendrik Hosang, Matthias Hein, and Bernt
  Schiele.
\newblock Weakly supervised object boundaries.
\newblock In {\em CVPR}, 2016.

\bibitem{scribble}
Di Lin, Jifeng Dai, Jiaya Jia, Kaiming He, and Jian Sun.
\newblock Scribblesup: Scribble-supervised convolutional networks for semantic
  segmentation.
\newblock In {\em CVPR}, 2016.

\bibitem{neverova}
Natalia Neverova and Iasonas Kokkinos.
\newblock Mass displacement networks.
\newblock In {\em BMVC}, 2018.

\bibitem{NewellYD16}
Alejandro Newell, Kaiyu Yang, and Jia Deng.
\newblock Stacked hourglass networks for human pose estimation.
\newblock In {\em {ECCV}}, 2016.

\bibitem{gpapanweak}
George Papandreou, Liang{-}Chieh Chen, Kevin Murphy, and Alan~L. Yuille.
\newblock Weakly- and semi-supervised learning of a {DCNN} for semantic image
  segmentation.
\newblock In {\em CVPR}, 2016.

\bibitem{PapandreouKS15}
George Papandreou, Iasonas Kokkinos, and Pierre{-}Andr{\'{e}} Savalle.
\newblock Modeling local and global deformations in deep learning: Epitomic
  convolution, multiple instance learning, and sliding window detection.
\newblock In {\em CVPR}, 2015.

\bibitem{dae}
Zhixin Shu, Mihir Sahasrabudhe, Riza~Alp G{\"{u}}ler, Dimitris Samaras, Nikos
  Paragios, and Iasonas Kokkinos.
\newblock Deforming autoencoders: Unsupervised disentangling of shape and
  appearance.
\newblock In {\em ECCV}, 2018.

\bibitem{Black00}
Hedvig Sidenbladh, Michael~J. Black, and David~J. Fleet.
\newblock Stochastic tracking of 3d human figures using 2d image motion.
\newblock In {\em ECCV}, 2000.

\bibitem{ThewlisBV17a}
James Thewlis, Hakan Bilen, and Andrea Vedaldi.
\newblock Unsupervised learning of object frames by dense equivariant image
  labelling.
\newblock In {\em NIPS}, 2017.

\bibitem{thewlis2017unsupervised}
James Thewlis, Hakan Bilen, and Andrea Vedaldi.
\newblock Unsupervised learning of object landmarks by factorized spatial
  embeddings.
\newblock In {\em ICCV}, 2017.

\bibitem{zhang2018unsupervised}
Yuting Zhang, Yijie Guo, Yixin Jin, Yijun Luo, Zhiyuan He, and Honglak Lee.
\newblock Unsupervised discovery of object landmarks as structural
  representations.
\newblock In {\em CVPR}, 2018.

\bibitem{ZhouKAHE16}
Tinghui Zhou, Philipp Kr{\"{a}}henb{\"{u}}hl, Mathieu Aubry, Qi{-}Xing Huang,
  and Alexei~A. Efros.
\newblock Learning dense correspondence via 3d-guided cycle consistency.
\newblock In {\em CVPR}, 2016.

\end{thebibliography}
\end{document}